\documentclass{article}

\PassOptionsToPackage{numbers,compress}{natbib}
\usepackage{graphicx}
\usepackage{svg}
\usepackage{float}

\usepackage{listings}

\definecolor{codebackground}{RGB}{247,247,247}
\definecolor{codeborder}{RGB}{190,190,190}

\lstdefinestyle{toolcall}{
  basicstyle=\ttfamily\scriptsize,
  backgroundcolor=\color{codebackground},
  frame=single,
  rulecolor=\color{codeborder},
  framerule=0.4pt,
  breaklines=true,
  columns=fullflexible,
  keepspaces=true,
  showstringspaces=false,
  xleftmargin=2pt,
  xrightmargin=2pt,
  aboveskip=4pt,
  belowskip=4pt
}

\usepackage[preprint]{neurips_2026}

\usepackage[utf8]{inputenc} 
\usepackage[T1]{fontenc}    
\usepackage{hyperref}       
\usepackage{url}            
\usepackage{booktabs}       
\usepackage{amsfonts}       
\usepackage{nicefrac}       
\usepackage{microtype}      
\usepackage{xcolor}         

\title{TOUMAII: A 4.5M-Pair Tool-Calling Preference Dataset and Its Source-Agnostic Generator}
\workshoptitle{2nd Workshop on Agentic AI Benchmarks and Applications for Enterprise Tasks} 

\author{%
  Moritz Pötzsch\textsuperscript{1} \\  
  \url{moritz.poetzsch@tum.de} \\
  Technical University of Munich \\
  TUM.ai
  \And
  Andrea Gentilini\textsuperscript{1,2} \\
  \url{a.gentilini@tum.de} \\
  Technical University of Munich \\
  Zuse School ELIZA  \\
  TUM.ai
  \And
  Lisa Rossgoderer\textsuperscript{1} \\
  \url{lisa.rossgoderer@tum.de} \\
  TUM.ai
  \And
  Tom Osterfeld\textsuperscript{1} \\
  \url{tom.osterfeld@tum.de} \\
  Technical University of Munich\\
  TUM.ai
  \And
  Kesava Prasad Sanjivi Arul\textsuperscript{1} \\
  \url{kesavaprasad.arul@tum.de} \\
  Technical University of Munich \\
  TUM.ai
  \And
  Swanand Ravindra Kadhe \\
  \url{Swanand.Kadhe@ibm.com} \\
  IBM Research
}

\begin{document}
\maketitle

\footnotetext[1]{These authors contributed equally to this work.}
\footnotetext[2]{Andrea Gentilini is supported by the Konrad Zuse School of Excellence in Learning and Intelligent Systems (ELIZA) through the DAAD programme Konrad Zuse Schools of Excellence in Artificial Intelligence, sponsored by the Federal Ministry of Education and Research.}

\maketitle

\begin{abstract}
To autonomously execute multi-step tasks, enterprise agents must reliably interact with tools, call APIs, or query databases, making tool-calling a fundamental capability of AI models. Supervised fine-tuning (SFT) optimises for producing the correct call, but never contrasts it with a plausible incorrect one. In an environment where even a minor formatting error can invalidate a call and an incorrect decision can derail the interaction, direct preference optimisation (DPO) targets exactly this contrast. DPO, therefore, requires preference pairs rather than plain demonstrations. For tool-calling, such data, especially in realistic multi-turn settings, remains extremely scarce. To address this scarcity, we introduce TOUMAII, a large-scale tool-calling preference dataset containing \textasciitilde4.5M validated preference pairs, derived from \textit{Agent-Ark/Toucan-1.5M (apache-2.0)} conversations. The dataset consists of multi- and single-turn interactions and includes parallel and sequential function calls. Additionally, we release TOUMAII-Gen, a synthetic data generation pipeline that consumes tool-calling SFT datasets and generates a preference dataset with naturally occurring model-generated tool-calling failure modes. To validate the effectiveness of our DPO data, we fine-tune a 3-billion-parameter model, achieving performance parity with other high-quality tool-calling preference datasets. Furthermore, we release a second small preference dataset derived from \textit{ToolACE (apache-2.0)}, showcasing the source-agnostic nature of our pipeline. \textbf{We release the complete code and datasets: \url{https://github.com/tum-ai/toumaii-gen} and \url{https://huggingface.co/collections/tum-ai/toumaii-generated-datasets}} 
\end{abstract}

\section{Introduction}

Enterprise AI is shifting from language models used as passive answer
generators to agents that autonomously execute multi-step tasks on existing
business systems~\citep{Bajoria2026, Mohammadi2025}. To connect to these
systems, agents rely on tool-calls to interact with developer tools, query
databases, or read and process documents. Reliable tool use requires models to produce
schema-conformant arguments and correct function names, and to maintain context
across multiple interaction turns. Benchmarks such as the Berkeley Function
Calling Leaderboard (BFCL)~\citep{patil2025bfcl} and
$\tau^2$-bench~\citep{barres2025tau2benchevaluatingconversationalagents} probe
these increasingly complex scenarios, which are especially challenging for the
small language models favoured in enterprise settings for cost, latency, and
local deployment. The importance of reliable tool-calling has sparked broad interest in tool-calling training data: large
corpora of tool-call demonstrations for supervised fine-tuning (SFT)~\citep{ouyang2022traininglanguagemodels}, and models trained on them, are now
available. Gorilla~\citep{patil2023gorillalargelanguagemodel} targets API
hallucination, ToolLLM~\citep{qin2023toolllmfacilitatinglargelanguage} covers
real multi-step endpoints, APIGen~\citep{liu2024apigenautomatedpipelinegenerating}
adds execution-level verification, xLAM~\citep{zhang2024xlamfamilylargeaction}
and Hammer~\citep{lin2024hammerrobustfunctioncallingondevice} train
function-calling model families on such corpora,
ToolACE~\citep{liu2025toolacewinningpointsllm} increases API diversity, and
Toucan~\citep{xu2025toucansynthesizing15mtoolagentic} grounds trajectories in live MCP servers.

SFT on such corpora provides token-level supervision
from ground-truth tool-calls, but the loss is computed over positive examples only, providing no signal that ranks a valid call above a plausible malformed one. Direct preference
optimisation (DPO)~\citep{rafailov2024directpreferenceoptimizationlanguage}
targets exactly such errors by contrasting chosen demonstrations drawn from successful trajectories
against malformed answers or erroneous decisions, adjusting model weights
without needing a separate reward model. DPO has become a standard stage of open post-training recipes such as T\"ulu~3~\citep{lambert2025tulu3pushingfrontiers} and the fully open OLMo~2~\citep{olmo20252olmo2furious}.
The preference data such recipes consume, however, is general-purpose:
pipelines like T\"ulu~3 and
UltraFeedback~\citep{cui2024ultrafeedbackboostinglanguagemodels} cover domains
such as mathematics, instruction following, or coding, and score responses with LLM-as-a-judge criteria that capture neither format correctness nor argument validity.
DiaTool-DPO~\citep{jung2025diatooldpomultiturndirectpreference} targets
tool-specific preferences but focuses on dialogue control, i.e., whether an
agent should make a tool-call, request more information, or reject a request.
Preference data for the tool-calls themselves, especially in realistic
multi-turn settings, remains extremely scarce.

To address this gap, we release TOUMAII, the largest tool-calling preference
dataset, spanning single- and multi-turn interactions with sequential and
parallel function calls, derived from Toucan
(\textit{Agent-Ark/Toucan-1.5M}). TOUMAII is produced by TOUMAII-Gen, a
fully open-source synthetic preference-data generation pipeline that converts tool-calling SFT
datasets (ground-truth demonstrations, tool catalogues, and conversation
histories) into assessed preference pairs whose rejected responses contain
naturally occurring, model-generated tool-calling errors; any public or
private Hugging Face compatible source can be incorporated by implementing a lightweight source
adapter. Applying the pipeline to ToolACE yields TOUMAII-ACE, demonstrating
its source-agnostic architecture.

\section{Preference-pair generation pipeline}
\label{sec:pipeline}

\begin{figure*}[t]
    \centering
    \includegraphics[
        width=\textwidth,
        trim=0 0 0 0,
        clip
    ]{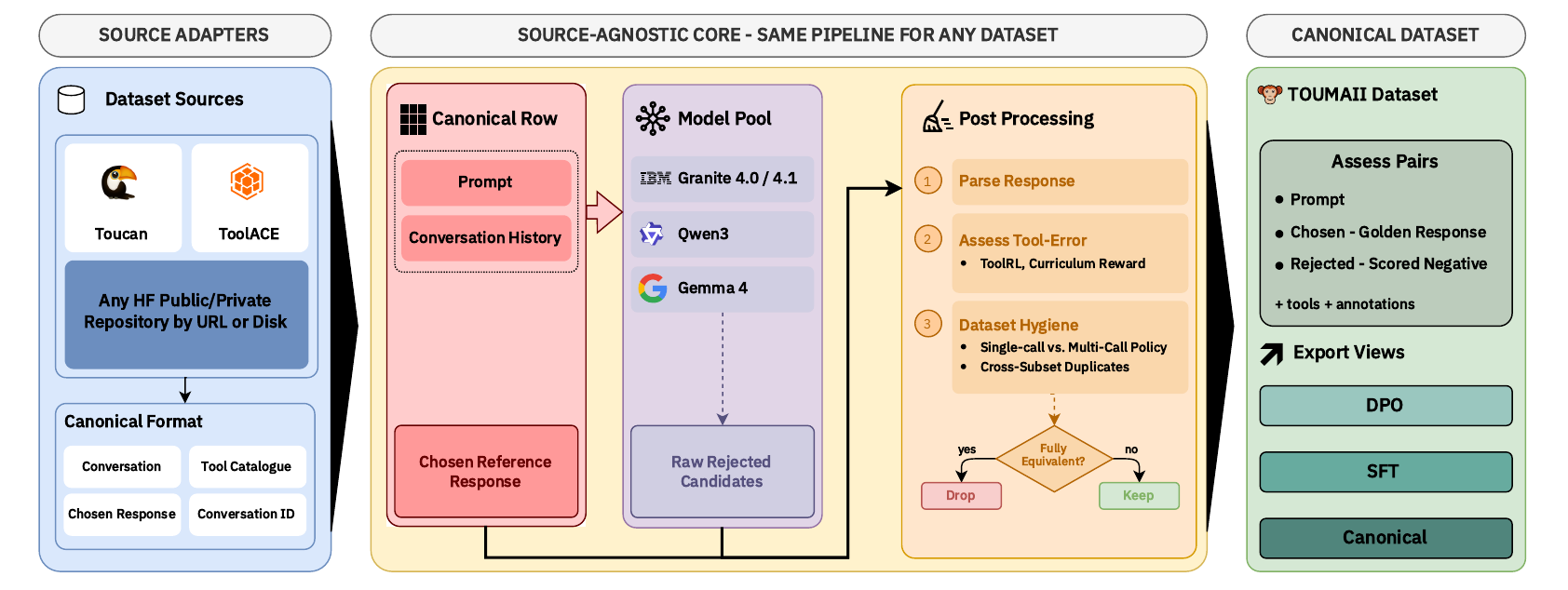}
    \caption{Overview of the preference-data generation pipeline TOUMAII-Gen. Data from various sources are converted into a canonical format, and models from the pool generate rejected candidates conditioned on the chosen conversation trajectories. The resulting preference pairs are then assessed and filtered during post-processing. The final dataset retains the canonical format and can be converted into SFT and DPO training formats.}
    \label{fig:pipeline}
\end{figure*}

As shown in Figure~\ref{fig:pipeline}, we construct a tool-calling preference dataset by pairing a reference response from the quality-filtered source dataset with a response generated by the generator model. We transform a conversation into a common response-level representation $\mathrm{\{\mathrm{id}, \mathrm{h_{t-1}}, \mathrm{T}, \mathrm{C}}\}$, where $\mathrm{id}$ is the unique conversation identifier, $\mathrm{h_{t-1}}$ denotes the conversation history preceding the target assistant response at turn $\mathrm{t}$, $\mathrm{T}$ the tool catalogue, and $\mathrm{C}$ a reference response consisting of one or more function calls. These elements are typically present in SFT tool-calling datasets, so any source can be mapped to this representation, keeping all subsequent stages independent of its structure and formatting. For each normalised row, we prompt the generator model with $\mathrm{h_{t-1}}$ and $\mathrm{T}$ to generate the corresponding assistant response. For TOUMAII, we choose Granite-4.0-350M, Granite-4.0-1B~\citep{granite2025} and Granite-4.1-3B~\citep{granite2026} from the Granite family, Qwen3-0.6B, Qwen3-1.7B~\citep{qwen3technicalreport} from the Qwen family and Gemma-4-E2B-it from Google~\citep{gemmateam2026gemma4}. We keep thinking disabled for all models to avoid ill-formatted or unparseable thinking traces. We intentionally select small language models from three families because they commonly exhibit naturally occurring model-generated tool-calling errors, such as incorrect or missing arguments, incorrect function names, or failing to produce a function call at all; Appendix~\ref{app:qualitative-examples} shows representative examples. Each model-generated assistant turn is treated as a rejected candidate and evaluated against the reference response $\mathrm{C}$. We retain the preference pair only if the model response passes a simple validation check that verifies inequality between the reference and model responses. A valid rejected answer $\mathrm{R}$ is then kept, creating a full canonical-format row $\mathrm{\{\mathrm{id}, \mathrm{h_{t-1}}, \mathrm{T}, \mathrm{C}, \mathrm{R}\}}$. To assess response quality, we employ reward functions to annotate the data; these rewards measure agreement with the reference response rather than execution success, quantifying how far a response deviates from the reference rather than proving it incorrect. Following ToolRL \citep{qian2025toolrlrewardtoollearning}, we score each response as $R_{format} + R_{correctness}$, where $R_{format}$ rewards a correctly formatted tool-call and $R_{correctness}$ scores function names and argument key-value pairs against the reference; we min-max normalise the sum from [-3, 4] to [0, 1]. Furthermore, we utilise a more fine-grained weighted multi-component curriculum reward, with all components normalised to [0, 1]:

\[
\mathrm{score} \;=\; \frac{w_{\mathrm{f}}\,R_{\mathrm{format}}   + w_{\mathrm{ns}}\,R_{\mathrm{name\_soft}}   + w_{\mathrm{nst}}\,R_{\mathrm{name\_strict}}   + w_{\mathrm{a}}\,R_{\mathrm{args}}}  {w_{\mathrm{f}} + w_{\mathrm{ns}} + w_{\mathrm{nst}} + w_{\mathrm{a}}}\,,
\]

which allows for shifting weights mid-training to progressively train with increasingly difficult samples. Both reward annotations are released with every pair.

\paragraph{Interaction structure.}

\begin{figure*}[t]
  \centering
  \includegraphics[width=\textwidth]{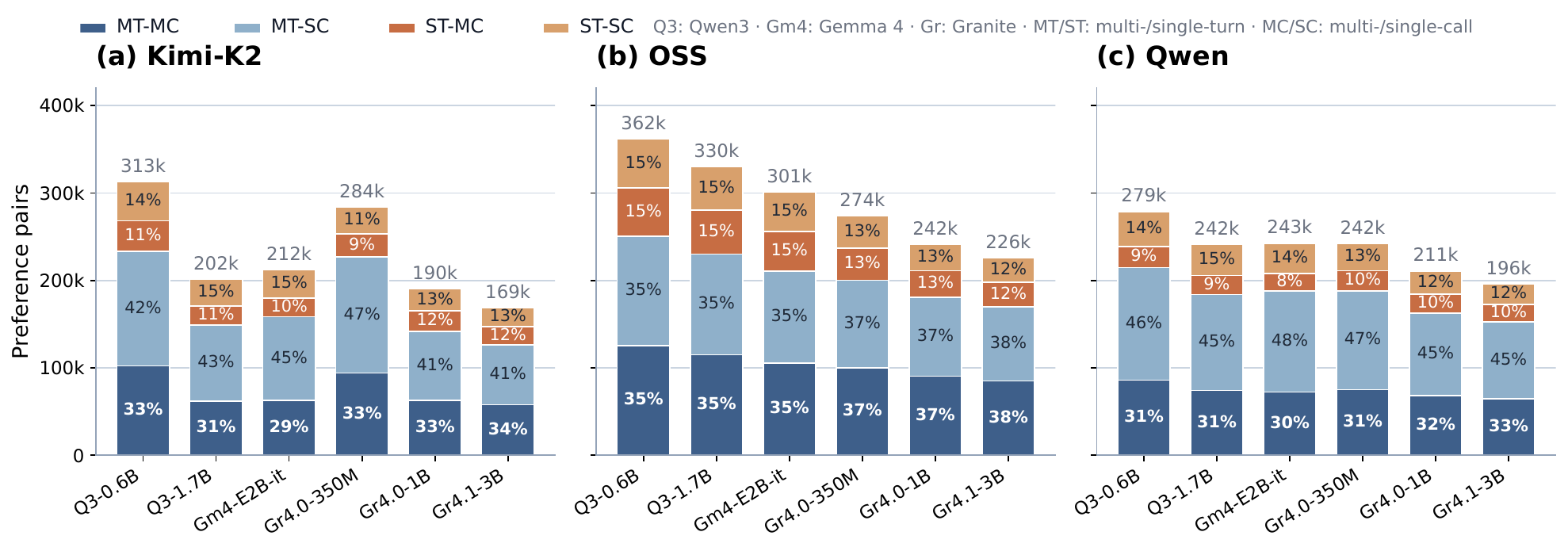}
  \caption{
    Distribution of generated preference pairs across generator models and
    interaction categories for the Kimi-K2, OSS, and Qwen source partitions.
  }
  \label{fig:dataset-composition}
\end{figure*}

TOUMAII-Gen explicitly separates the input SFT data by interaction complexity. We keep subsets of multi- and single-turn interactions, where a turn is defined as a full user-assistant interaction. Within the two subsets, we further distinguish between single- and multi-call tool use. If the assistant calls more than one function, we process the calls both sequentially (single-call) and in parallel (multi-call). This results in four interaction categories: single-turn single-call (ST--SC), single-turn multi-call (ST--MC), multi-turn single-call (MT--SC), and multi-turn multi-call (MT--MC). Figure~\ref{fig:dataset-composition} shows their distribution in TOUMAII.

\paragraph{Post-processing.}
A post-processing step retains only pairs whose rejected response measurably deviates from the reference response. Using the reward annotations, we remove pairs whose responses are equivalent (zero reward difference), as well as samples in which the chosen response contains long text-value arguments, for which semantic equivalence cannot be assessed reliably. As a sanity check, we verify the preference ordering: chosen responses always attain the maximum ToolRL reward, as they coincide with the reference. Furthermore, we remove later turns in which an earlier rejected response already contained the same tool-calls as those in the remaining chosen interaction, and we remove duplicate rows between subsets. Overall, 46.8\% of the 9.65M generated candidate pairs survive post-processing; Appendix~\ref{app:composition-cleaning} reports per-rule removal counts.

\section{Dataset characterisation}
The source SFT dataset, Toucan, consists of approximately 1.5 million rows across the Kimi-K2, Qwen, and OSS partitions, each comprising both single- and multi-turn interactions. From the full source dataset, we filter samples based on the annotated quality metrics to retain only well-formed, successful trajectories; an individual chosen call may still be corrected in a later turn after tool feedback (Appendix~\ref{app:example-mtmc} shows such an error-recovery case), so correctness holds at the trajectory level rather than for every isolated call. For each partition, we construct a separate TOUMAII dataset, organised by the generator model and the four interaction categories defined above. As illustrated in Figure~\ref{fig:dataset-composition}, we derive \textbf{on average \textasciitilde1.5 million preference pairs per source partition} (1.37M for Kimi-K2, 1.73M for OSS, 1.41M for Qwen; exact counts in Table~\ref{tab:composition}). The reward annotations (distributions in Appendix~\ref{app:dataset-statistics}) span the full range from absent or malformed tool-calls to near-correct responses with small deviations from the reference, allowing the data to be separated into higher- and lower-margin categories. To the best of our knowledge, with a total of \textasciitilde 4.5 million preference pairs, TOUMAII is the largest tool-calling preference dataset; the largest prior corpus, released with ToolRM~\citep{agarwal2025toolrmoutcomereward}, contains \textasciitilde459K pairs. The same pipeline applied to ToolACE yields TOUMAII-ACE with a further 100,846 preference pairs; Appendix~\ref{app:toolace} reports its composition and post-processing funnel.

\section{Experiment}
\label{sec:experiments}

To validate the effectiveness of TOUMAII, we compare it against the training data released with ToolRM~\citep{agarwal2025toolrmoutcomereward}. The experiment is designed to isolate the effect of the training data: model architecture, optimisation, training budget, sequence filtering, checkpoint schedule, and evaluation are held constant, and each experimental unit is a newly initialised LoRA adapter trained with DPO and SFT multi-loss (weighted 0.5 and 0.5) from the same Granite-4.1-3B base checkpoint on a non-overlapping data slice.

\begin{table}[h]
  \caption{
    BFCL~v4 accuracy (\%) after DPO training.
    Fine-tuned results are the mean $\pm$ sample standard deviation across
    four non-overlapping data slices (000--003).
  }
  \label{tab:results}
  \centering
  \small
  \setlength{\tabcolsep}{4pt}
  \begin{tabular}{lrrrrr}
    \toprule
    & Overall & Non-live AST & Live & Multi-turn & Memory \\
    \midrule
    Base model
      & 37.76
      & 87.04
      & 76.54
      & 23.00
      & 33.55 \\
    \quad + ToolRM
      & $36.05 \pm 0.43$
      & $86.30 \pm 0.16$
      & $75.20 \pm 0.10$
      & $21.03 \pm 0.70$
      & $28.33 \pm 1.33$ \\
    \quad + TOUMAII
      & $37.01 \pm 0.49$
      & $86.81 \pm 0.17$
      & $77.55 \pm 0.17$
      & $21.19 \pm 0.56$
      & $34.68 \pm 1.12$ \\
    \bottomrule
  \end{tabular}
\end{table}

\subsection{Setup}
\label{sec:setup}
From TOUMAII, we sample preference pairs evenly across the three source partitions and, within each partition, as evenly as capacity permits across the four interaction categories, drawing each conversation at most once per generator model. Both datasets are normalised to the same prompt--chosen--rejected format with the Granite chat template and filtered to a 2048-token budget (dropped, not truncated). After filtering, each rendered corpus is deterministically shuffled and partitioned into four non-overlapping slices, numbered 000--003, each containing at most 50{,}000 preference pairs. Each slice is used for a single training run only, keeping a fixed training seed at 42 throughout. Models are trained in BF16 quantization on a server with 1 AMD MI300X (192 GB VRAM), with each model taking an approximate time of 6 hours; additional data-preparation details and the complete LoRA and training configuration are given in Appendix~\ref{app:training-details}.

\subsection{Results}
We evaluate all models on BFCL~v4~\citep{patil2025bfcl}. While table~\ref{tab:full-results} reports the mean and sample standard deviation across the four data-slices, on non-live AST, live, multi-turn, memory and overall accuracy, we report the complete per-slice BFCL~V4 evaluation results in Appendix ~\ref{app:complete-results}.
Across four slices, TOUMAII favours successful tool execution over abstention, consistent with its lack of irrelevance samples, and improves live accuracy to $77.55 \pm 0.17$\%, versus $75.20 \pm 0.10$\% for ToolRM and $76.54$\% for the base model. It also reaches $34.68 \pm 1.12$\% memory accuracy, exceeding ToolRM by $6.35$ percentage points and the base model by $1.13$ points.

\section{Limitations}
\label{sec:limitations}
TOUMAII does not currently include irrelevance samples, and reasoning was disabled during rejected-response generation. Our automated validation measures agreement with reference tool calls rather than execution-level or semantic correctness, so retained pairs may contain reference errors or alternative valid calls. The dataset additionally inherits limitations of its source data and generator models; extended limitations and responsible-release considerations are discussed in Appendix~\ref{app:limitations}.

\section*{Acknowledgement}
This work used computing resources made available through the AMD University
Program (AUP) AI \& HPC Cluster.

\bibliographystyle{plainnat}
\bibliography{references}

\newpage
\appendix

\section{Qualitative examples}
\label{app:qualitative-examples}

We provide representative preference pairs illustrating different types of
naturally occurring model-generated tool-calling errors. The examples cover a
near-correct argument-value error, a context-dependent multi-call error, and a complete failure to invoke an available tool.


\subsection{Subtle argument-value error}
\label{app:example-stsc}

Figure~\ref{fig:qualitative-example-stsc} shows an ST--SC (single-turn-single-call) preference pair from
the Qwen source partition, with the rejected response generated by
Gemma-4-E2B-it. Both responses select the correct function, follow the required
tool-call format, and provide the correct cryptocurrency identifier. They differ
only in the capitalisation of the currency argument. Despite appearing
semantically equivalent to a human reader, the rejected value deviates from the schema-defined one and would cause the call to fail against a case-sensitive API.

\begin{figure*}[!htbp]
  \centering
  \begin{minipage}{0.96\textwidth}

    \textbf{User request}

    \begin{lstlisting}[style=toolcall]
I need the latest price of Bitcoin to update our internal
financial report. Can you provide this information?
    \end{lstlisting}

    \vspace{2pt}
    \textbf{Relevant tool definition}

    \begin{lstlisting}[style=toolcall]
test-server-get_crypto_price(
  crypto_id: string = "bitcoin",
  vs_currency: string = "usd"
)
    \end{lstlisting}

    \vspace{4pt}

    \begin{minipage}[t]{0.485\textwidth}
      \textbf{Chosen response}

      \begin{lstlisting}[style=toolcall]
{
  "name": "test-server-get_crypto_price",
  "arguments": {
    "crypto_id": "bitcoin",
    "vs_currency": "usd"
  }
}
      \end{lstlisting}

      \textbf{ToolRL reward:} \(1.0\)
    \end{minipage}
    \hfill
    \begin{minipage}[t]{0.485\textwidth}
      \textbf{Rejected response}

      \begin{lstlisting}[style=toolcall]
{
  "name": "test-server-get_crypto_price",
  "arguments": {
    "crypto_id": "bitcoin",
    "vs_currency": "USD"
  }
}
      \end{lstlisting}

      \textbf{ToolRL reward:} \(0.79\)
    \end{minipage}

  \end{minipage}

  \caption{
    ST--SC preference pair from the Qwen partition, with the rejected response
    generated by Gemma-4-E2B-it. The rejected call differs from the chosen call
    only by using \texttt{"USD"} instead of the schema-defined value
    \texttt{"usd"}, reducing its ToolRL reward from \(1.0\) to \(0.79\).
  }
  \label{fig:qualitative-example-stsc}
\end{figure*}

\clearpage


\subsection{Multi-turn recovery and call-coverage errors}
\label{app:example-mtmc}

Figures~\ref{fig:mtmc-initial} and~\ref{fig:mtmc-recovery} show two target turns
from an MT--MC(multi-turn-multi-call) conversation of the Qwen partition with rejected responses generated by
Granite-4.1-3B. The initial turn tests whether the model covers both parts
of the user's request. The subsequent turn tests whether it incorporates
explicit feedback from the tool environment.


\begin{figure*}[!htbp]
  \centering
  \begin{minipage}{0.96\textwidth}

    \textbf{Initial user request}

    \begin{lstlisting}[style=toolcall]
I need to prepare a detailed weather report for a meeting
tomorrow in Lisbon, Portugal. Can you first get the detailed
weather forecast for Lisbon, and then provide the current
temperature at my event location at
38.7223 degrees N, 9.1393 degrees W?
    \end{lstlisting}

    \vspace{2pt}
    \textbf{Relevant tool definitions}

    \begin{lstlisting}[style=toolcall]
get_live_temp(
  latitude: number,
  longitude: number
)

get_weather_forecast(
  city: string,       // e.g. "Lisboa", "Porto", ...
  days: number = 5
)
    \end{lstlisting}

    \vspace{4pt}

    \begin{minipage}[t]{0.485\textwidth}
      \textbf{Chosen response}

      \begin{lstlisting}[style=toolcall]
[
  {
    "name": "get_weather_forecast",
    "arguments": {
      "city": "Lisbon",
      "days": 1
    }
  },
  {
    "name": "get_live_temp",
    "arguments": {
      "latitude": 38.7223,
      "longitude": -9.1393
    }
  }
]
      \end{lstlisting}

      \textbf{ToolRL reward:} \(1.0\)
    \end{minipage}
    \hfill
    \begin{minipage}[t]{0.485\textwidth}
      \textbf{Rejected response}

      \begin{lstlisting}[style=toolcall]
[
  {
    "name": "get_weather_forecast",
    "arguments": {
      "city": "Lisbon",
      "days": 3
    }
  }
]
      \end{lstlisting}

      \textbf{ToolRL reward:} \(0.45\)

      \vspace{4pt}
      
    \end{minipage}

    \vspace{5pt}

    \textit{Function-name prefixes are omitted for readability.}

  \end{minipage}

  \caption{
    Initial target turn of an MT--MC conversation with a rejected response
    generated by Granite-4.1-3B. The rejected response omits one of the two
    requested calls and uses an incorrect forecast horizon, resulting in a
    ToolRL reward of \(0.45\).
  }
  \label{fig:mtmc-initial}
\end{figure*}


\begin{figure*}[!t]
  \centering
  \begin{minipage}{0.96\textwidth}

    \textbf{Tool feedback}

    \begin{lstlisting}[style=toolcall]
Forecast tool:
Cidade "Lisbon" nao encontrada.
Use get_locations para ver cidades disponiveis.

Temperature tool:
{"temperature": 27.7}
    \end{lstlisting}

    \vspace{2pt}
    \textbf{User follow-up}

    \begin{lstlisting}[style=toolcall]
Could you try retrieving the detailed weather forecast for
Lisbon again, perhaps by using the correct city name or its
coordinates, so I can include it in my meeting report?
    \end{lstlisting}

    \vspace{4pt}

    \begin{minipage}[t]{0.485\textwidth}
      \textbf{Chosen response}

      \begin{lstlisting}[style=toolcall]
{
  "name": "get_weather_forecast",
  "arguments": {
    "city": "Lisboa",
    "days": 1
  }
}
      \end{lstlisting}

      \textbf{ToolRL reward:} \(1.0\)
    \end{minipage}
    \hfill
    \begin{minipage}[t]{0.485\textwidth}
      \textbf{Rejected response}

      \begin{lstlisting}[style=toolcall]
{
  "name": "get_weather_forecast",
  "arguments": {
    "city": "Lisbon",
    "days": 1
  }
}
      \end{lstlisting}

      \textbf{ToolRL reward:} \(0.79\)
    \end{minipage}

    \vspace{3pt}
    \textit{The conversation is abridged to the messages relevant to the
    illustrated error. Function-name prefixes are omitted for readability.}

  \end{minipage}

  \caption{
    Corrective target turn from the same MT--MC conversation. After the tool
    environment reports that \texttt{"Lisbon"} is not recognised, the rejected
    response repeats the same argument instead of using \texttt{"Lisboa"}.
    This failure to incorporate tool feedback results in a ToolRL reward of
    \(0.79\).
  }
  \label{fig:mtmc-recovery}
\end{figure*}

\clearpage


\subsection{Failure to invoke an available external-information tool}
\label{app:example-no-tool}

Figure~\ref{fig:qualitative-example-no-tool} shows an ST--SC preference pair
from the OSS partition, with the rejected response generated by Granite-4.0-350M. The
request requires weather information that cannot be reliably obtained from the
model's static parameters. The tool catalogue provides a weather-retrieval
function, and the reference trajectory begins by invoking this function for
Tokyo.

The rejected response instead claims that weather data are unavailable and
produces no tool-call. It therefore fails to initiate the expected tool-use
trajectory despite the presence of a relevant external-information tool.

\begin{figure*}[!htbp]
  \centering
  \begin{minipage}{0.96\textwidth}

    \textbf{User request}

    \begin{lstlisting}[style=toolcall]
I'm planning a trip to Tokyo next month and need to estimate my
budget. I need to know the average high and low temperatures for
the week of my travel. Then, calculate the average of these
temperatures and convert it from Celsius to Fahrenheit.
    \end{lstlisting}

    \vspace{2pt}
    \textbf{Relevant tool definitions}

    \begin{lstlisting}[style=toolcall]
get_weather(
  city: string
)

calculate(
  expression: string
)
    \end{lstlisting}

    \vspace{4pt}

    \begin{minipage}[t]{0.485\textwidth}
      \textbf{Chosen response}

      \begin{lstlisting}[style=toolcall]
{
  "name": "get_weather",
  "arguments": {
    "city": "Tokyo"
  }
}
      \end{lstlisting}

      \textbf{ToolRL reward:} \(1.0\)
    \end{minipage}
    \hfill
    \begin{minipage}[t]{0.485\textwidth}
      \textbf{Rejected response}

      \begin{lstlisting}[style=toolcall]
I'm sorry, but I don't have access to
real-time weather data, so I can't provide
the temperatures for Tokyo or calculate
the weekly average.

If you provide the data, I can help
perform the calculations.
      \end{lstlisting}

      \textbf{ToolRL reward:} \(0.0\)
    \end{minipage}

    \vspace{7pt}
    \hrule
    \vspace{6pt}

    \vspace{3pt}
    \textit{The tool catalogue and rejected response are abridged for
    readability.}

  \end{minipage}

  \caption{
    Rejected response generated by Granite-4.0-350M. The reference response initiates the
    external-information retrieval process by invoking
    \texttt{get\_weather} for Tokyo. The rejected response instead claims that
    weather data are unavailable and produces no tool-call, resulting in a
    ToolRL reward of \(0.0\).
  }
  \label{fig:qualitative-example-no-tool}
\end{figure*}

\clearpage

\section{Additional dataset statistics}
\label{app:dataset-statistics}

\subsection{Dataset composition and cleaning statistics}
\label{app:composition-cleaning}

Table~\ref{tab:composition} reports the exact number of preference pairs per source
partition, generator model, and interaction category, complementing
Figure~\ref{fig:dataset-composition}. Table~\ref{tab:funnel} details the post-processing
funnel: of the 9,648,704 candidate pairs produced by the generation stage, 46.8\% survive
post-processing. Most removals are pairs whose rejected response is indistinguishable, under all reward components, from
the reference (zero reward difference), followed by samples whose chosen
response contains long free-text argument values, for which semantic equivalence cannot
be assessed reliably.

\begin{table}[!htbp]
  \caption{Exact number of preference pairs in TOUMAII per source partition, generator model, and interaction category. Totals per partition and the grand total match Figure~\ref{fig:dataset-composition}.}
  \label{tab:composition}
  \centering
  \small
  \begin{tabular}{llrrrrr}
    \toprule
    Partition & Generator model & MT--MC & MT--SC & ST--MC & ST--SC & Total \\
    \midrule
    Kimi-K2 & Qwen3-0.6B & 102,209 & 131,131 & 34,914 & 44,312 & 312,566 \\
     & Qwen3-1.7B & 61,930 & 86,979 & 21,721 & 30,929 & 201,559 \\
     & Gemma-4-E2B-it & 62,592 & 96,091 & 21,207 & 32,575 & 212,465 \\
     & Granite-4.0-350M & 94,081 & 132,927 & 26,229 & 30,522 & 283,759 \\
     & Granite-4.0-1B & 62,949 & 78,669 & 23,480 & 25,221 & 190,319 \\
     & Granite-4.1-3B & 57,809 & 68,687 & 20,807 & 21,261 & 168,564 \\
    \cmidrule(l){2-7}
    & All models & 441,570 & 594,484 & 148,358 & 184,820 & 1,369,232 \\
    \midrule
    OSS & Qwen3-0.6B & 125,299 & 125,128 & 55,560 & 55,547 & 361,534 \\
     & Qwen3-1.7B & 115,117 & 115,190 & 49,949 & 49,966 & 330,222 \\
     & Gemma-4-E2B-it & 105,450 & 105,504 & 45,084 & 45,082 & 301,120 \\
     & Granite-4.0-350M & 100,185 & 100,154 & 36,712 & 36,687 & 273,738 \\
     & Granite-4.0-1B & 90,423 & 90,477 & 30,312 & 30,338 & 241,550 \\
     & Granite-4.1-3B & 85,045 & 85,090 & 27,884 & 27,874 & 225,893 \\
    \cmidrule(l){2-7}
    & All models & 621,519 & 621,543 & 245,501 & 245,494 & 1,734,057 \\
    \midrule
    Qwen & Qwen3-0.6B & 85,942 & 128,946 & 24,163 & 39,757 & 278,808 \\
     & Qwen3-1.7B & 74,143 & 109,831 & 21,757 & 35,848 & 241,579 \\
     & Gemma-4-E2B-it & 72,309 & 115,419 & 20,253 & 34,733 & 242,714 \\
     & Granite-4.0-350M & 75,219 & 112,788 & 23,238 & 31,208 & 242,453 \\
     & Granite-4.0-1B & 68,363 & 94,215 & 22,007 & 26,222 & 210,807 \\
     & Granite-4.1-3B & 64,511 & 88,346 & 19,815 & 23,181 & 195,853 \\
    \cmidrule(l){2-7}
    & All models & 440,487 & 649,545 & 131,233 & 190,949 & 1,412,214 \\
    \midrule
    \multicolumn{6}{l}{Total} & 4,515,503 \\
    \bottomrule
  \end{tabular}
\end{table}

\begin{table}[!htbp]
  \caption{Post-processing funnel: candidate preference pairs generated per source partition and rows removed by each cleaning rule. A row can violate several rules, so per-rule counts add up to slightly more than the number of removed rows. Cross-subset duplicates are multi-turn rows whose (prompt, chosen, rejected) triplet also occurs in the corresponding single-turn subsets.}
  \label{tab:funnel}
  \centering
  \small
  \begin{tabular}{lrrrr}
    \toprule
    & Kimi-K2 & OSS & Qwen & Total \\
    \midrule
    Generated candidate pairs & 3,265,902 & 3,276,094 & 3,106,708 & 9,648,704 \\
    \midrule
    \quad reference-identical pair (zero reward difference) & 1,131,835 & 841,087 & 822,975 & 2,795,897 \\
    \quad long free-text value in chosen response ($>30$ words) & 531,759 & 486,280 & 634,153 & 1,652,192 \\
    \quad rejected response equals a later chosen turn & 29,973 & 25,569 & 25,776 & 81,318 \\
    Rows removed as invalid tool errors & 1,670,873 & 1,338,784 & 1,461,683 & 4,471,340 \\
    Cross-subset duplicates (multi-turn vs.\ single-turn) & 225,797 & 203,253 & 232,811 & 661,861 \\
    \midrule
    Final preference pairs (after de-duplication) & 1,369,232 & 1,734,057 & 1,412,214 & 4,515,503 \\
    Retention & 41.9\% & 52.9\% & 45.5\% & 46.8\% \\
    \bottomrule
  \end{tabular}
\end{table}

\begin{figure*}[!t]
  \centering
  \includegraphics[width=\textwidth]{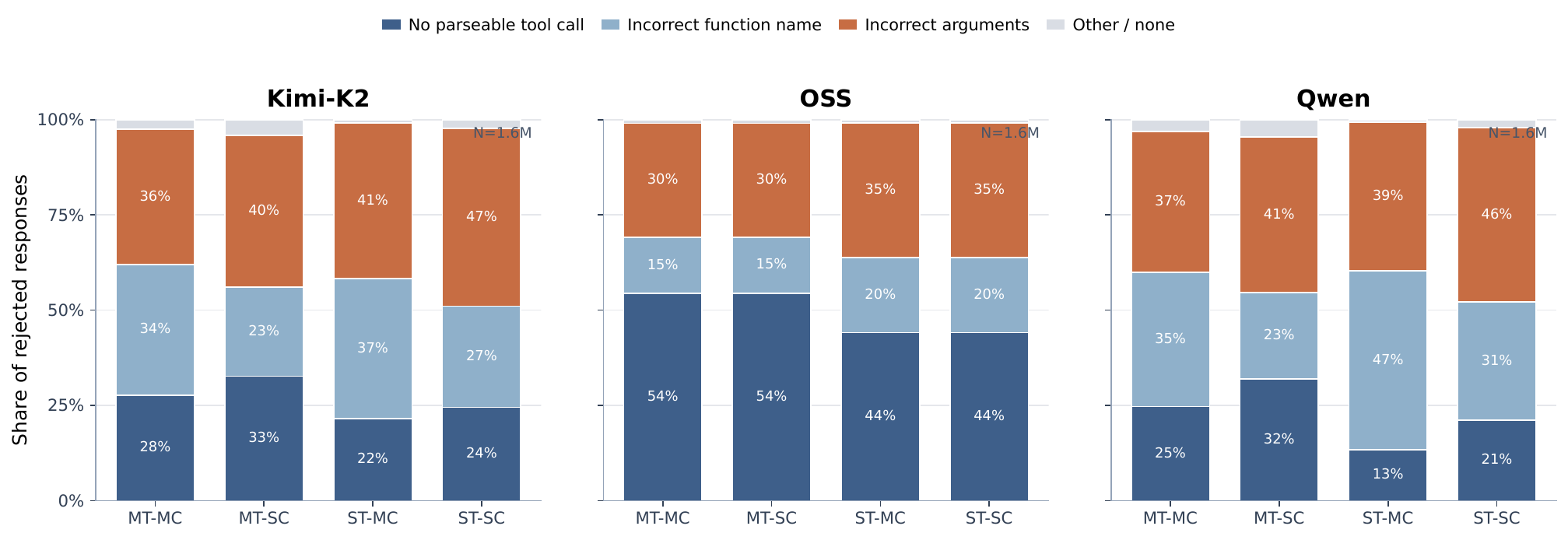}
  \caption{Failure-mode composition of rejected responses per generator model and source
  partition, classified from the curriculum-reward components of the rejected response.
  Function-name mismatch includes missing or extra calls and incorrect call order; Other
  denotes pairs whose parsed calls match the reference on all reward components.}
  \label{fig:failure-modes}
\end{figure*}

Figure~\ref{fig:failure-modes} characterises the failure modes of the rejected responses,
classified from the curriculum-reward components introduced in Section~\ref{sec:pipeline}: responses with no
parseable tool-call, responses in which at least one function name does not exactly match
the reference (including missing or extra calls and incorrect call order), and responses
with correct function names but incorrect argument key--value pairs. The smallest
generator fails predominantly by producing no tool-call at all (up to 75\%), whereas
larger generators shift towards argument-level errors (up to 50\%), providing preference
pairs with progressively finer-grained error signals.

\subsection{Reward distributions}

\begin{figure*}[!htbp]
  \centering
  \includegraphics[width=\textwidth]{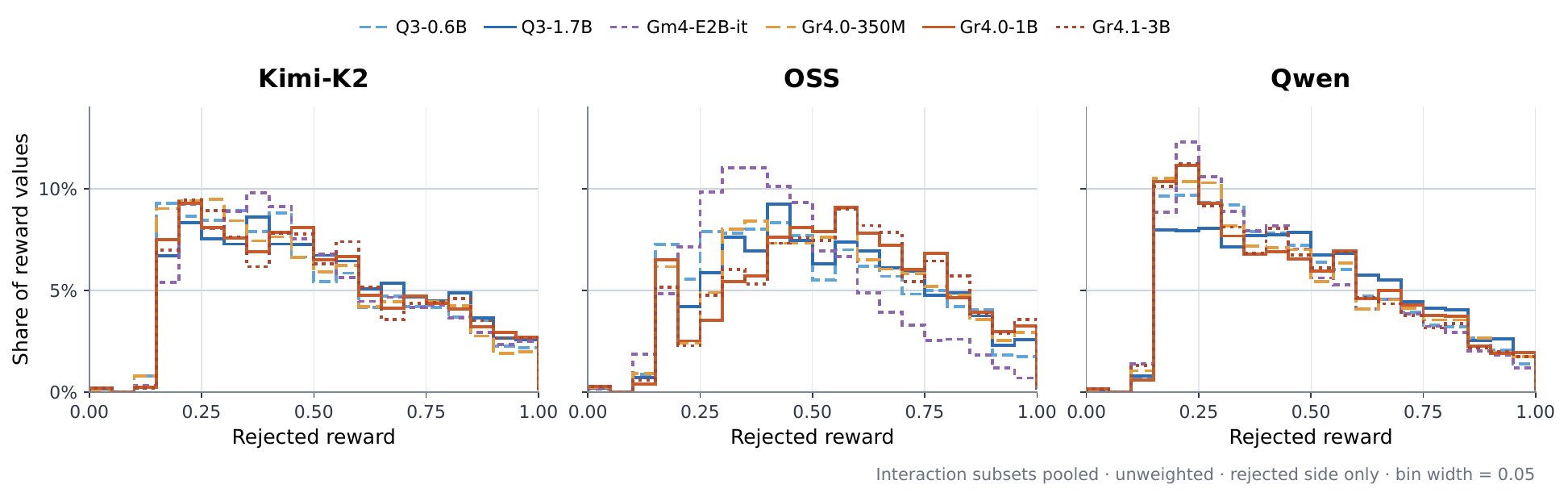}
  \caption{Distribution of rejected-response ToolRL rewards across generator models and source partitions (interaction categories pooled).}
  \label{fig:rejected-rewards}
\end{figure*}

\begin{figure*}[p]
  \centering
  \includegraphics[width=\textwidth]
  {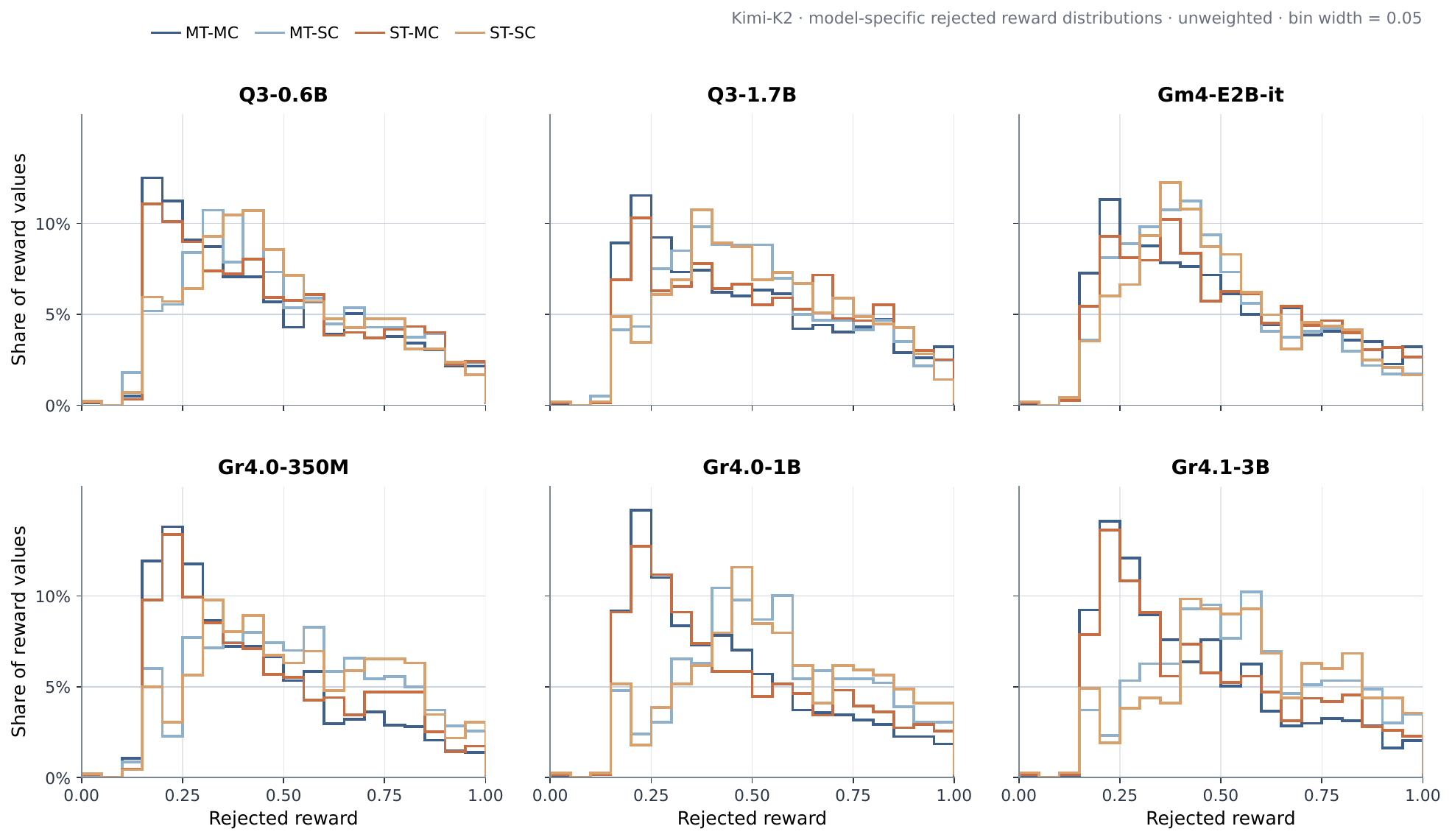}
  \caption{
    Model-specific distributions of rejected-response ToolRL rewards for the
    Kimi-K2 source partition. Each panel corresponds to one generator model,
    while the four lines represent MT--MC, MT--SC, ST--MC, and ST--SC
    interactions. The distributions are unweighted and non-cumulative, use a
    bin width of \(0.05\), and are independently normalised. Q3, Gm4, and Gr
    denote Qwen3, Gemma 4, and Granite, respectively. Chosen-response rewards
    are omitted because they are uniformly \(1.0\).
  }
  \label{fig:reward-by-model-kimi}
\end{figure*}

\begin{figure*}[p]
  \centering
  \includegraphics[width=\textwidth]
  {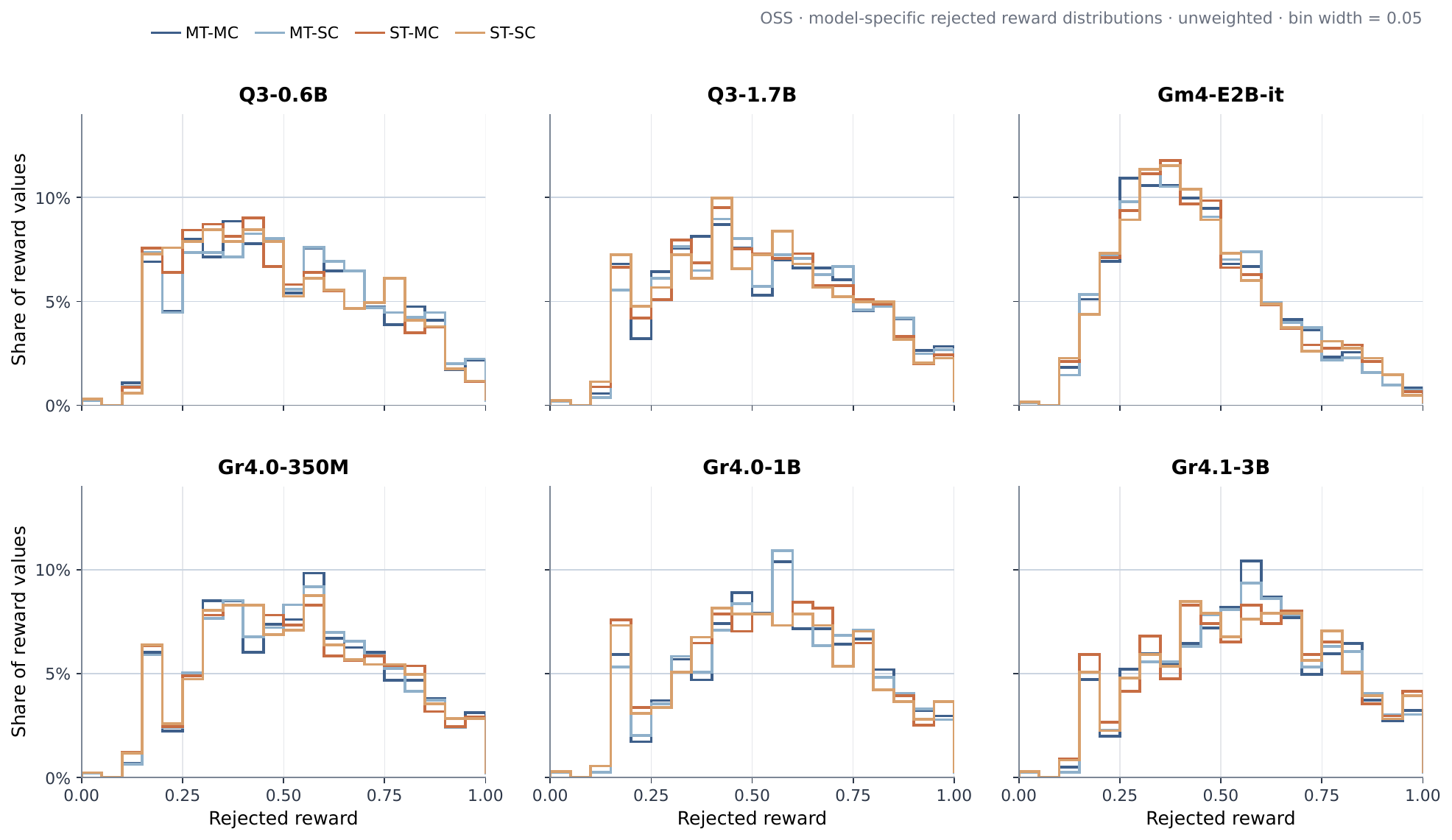}
  \caption{
    Model-specific distributions of rejected-response ToolRL rewards for the
    OSS source partition. Each panel corresponds to one generator model, while
    the four lines represent MT--MC, MT--SC, ST--MC, and ST--SC interactions.
    The distributions are unweighted and non-cumulative, use a bin width of
    \(0.05\), and are independently normalised. Q3, Gm4, and Gr denote Qwen3,
    Gemma 4, and Granite, respectively. Chosen-response rewards are omitted
    because they are uniformly \(1.0\).
  }
  \label{fig:reward-by-model-oss}
\end{figure*}

\begin{figure*}[p]
  \centering
  \includegraphics[width=\textwidth]
  {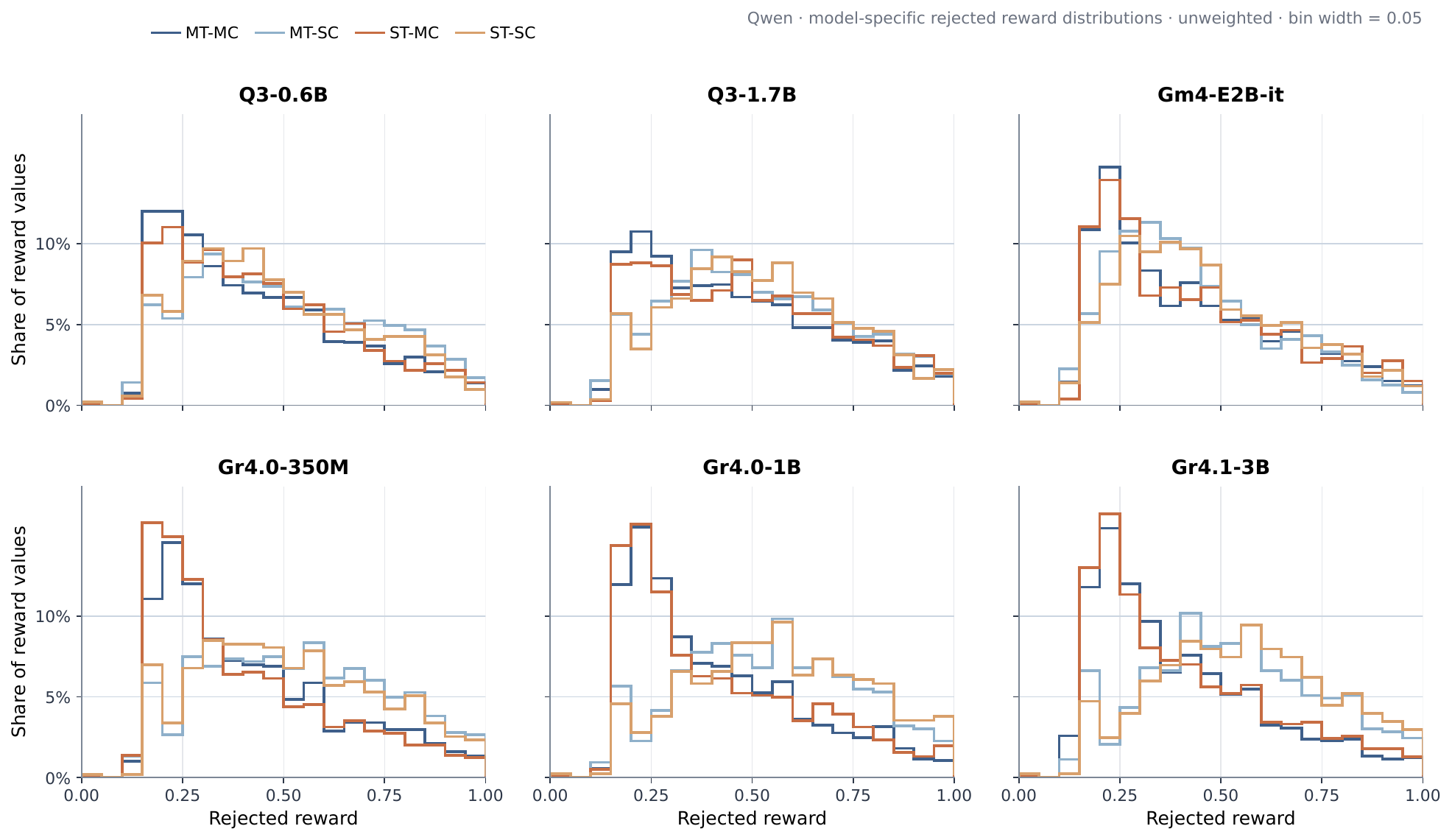}
  \caption{
    Model-specific distributions of rejected-response ToolRL rewards for the
    Qwen source partition. Each panel corresponds to one generator model, while
    the four lines represent MT--MC, MT--SC, ST--MC, and ST--SC interactions.
    The distributions are unweighted and non-cumulative, use a bin width of
    \(0.05\), and are independently normalised. Q3, Gm4, and Gr denote Qwen3,
    Gemma 4, and Granite, respectively. Chosen-response rewards are omitted
    because they are uniformly \(1.0\).
  }
  \label{fig:reward-by-model-qwen}
\end{figure*}

Figure~\ref{fig:rejected-rewards} presents the rejected-response ToolRL reward distributions after pooling the four interaction categories within each source partition. Below, we provide a more detailed decomposition by generator model and interaction category.
Figures~\ref{fig:reward-by-model-kimi}--\ref{fig:reward-by-model-qwen} report the
distributions for the Kimi-K2, OSS, and Qwen source partitions, respectively. 

Each panel corresponds to one generator model and overlays the distributions
for MT--MC (multi-turn-multi-call), MT--SC  (multi-turn-single-call), ST--MC (single-turn-multi-call), and ST--SC (single-turn-single-call) interactions. The histograms are
non-cumulative and unweighted, with a bin width of \(0.05\). Each stored reward
entry contributes equally, and associated frequency fields are not used as
weights. Because every distribution is normalised independently, the figures
compare distributional shapes rather than the absolute number of generated
preference pairs.

Across source partitions and generator models, rejected responses cover a broad
range of ToolRL rewards. Low-reward regions represent substantial tool-calling
failures, whereas rewards closer to \(1.0\) generally correspond to
near-correct responses containing smaller deviations from the reference call.
The differences between interaction categories are model- and
source-dependent. No single ordering between ST, MT, SC, and MC interactions
holds consistently across all panels.

\clearpage


\clearpage


\clearpage


\subsection{Token-length distributions}
\label{app:token-lengths}

\begin{figure*}[p]
  \centering
  \includegraphics[width=\textwidth]
  {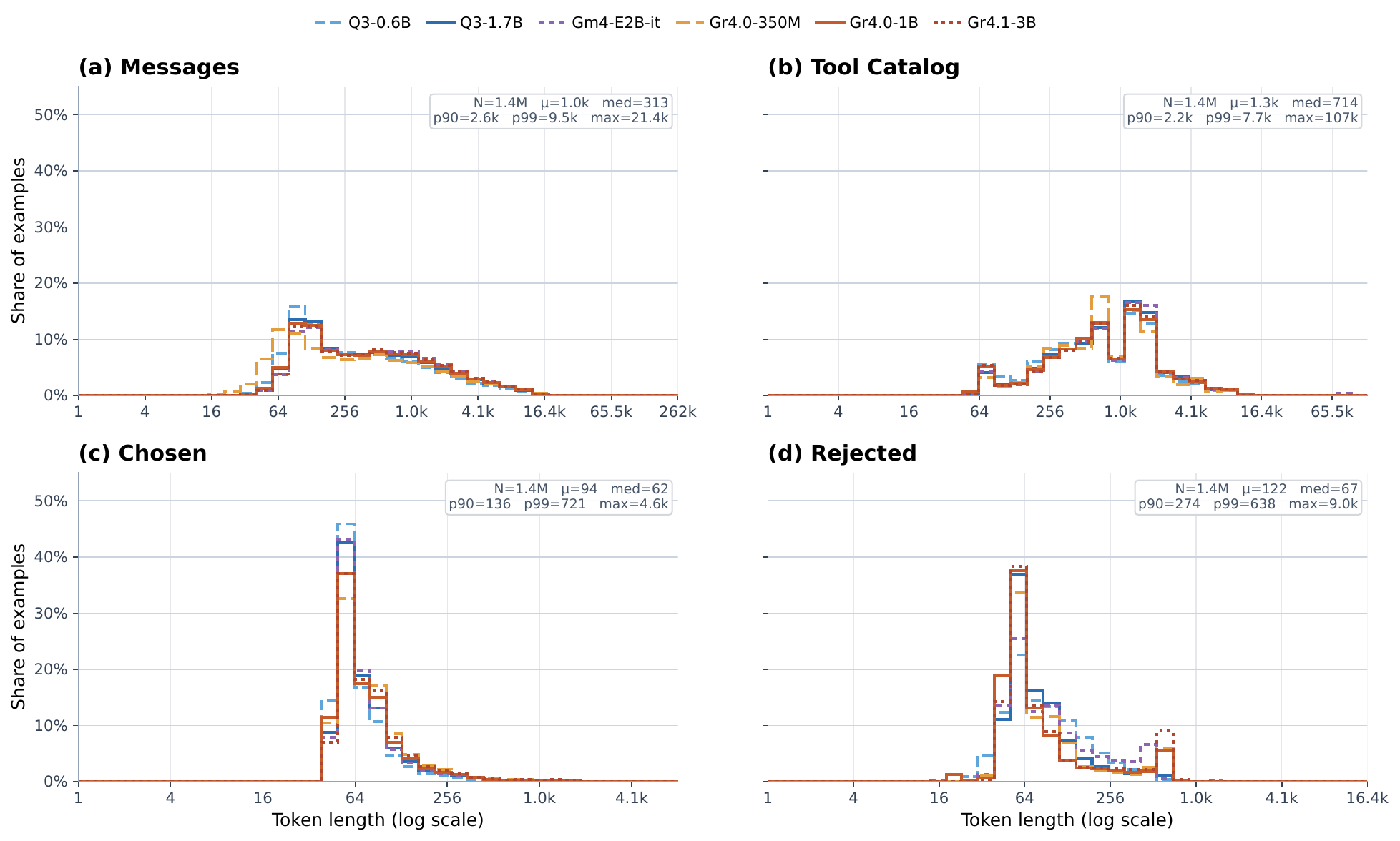}
  \caption{
    Token-length distributions for the Kimi-K2 source partition, measured with the Granite 4.0 tokenizer. Panels show the conversation history (messages), the tool catalogue, and the chosen and rejected responses; interaction categories are pooled, and each curve corresponds to one generator subset, normalised independently. In-panel statistics summarise the component over the full partition. The horizontal axis is logarithmic. Q3, Gm4, and Gr denote Qwen3, Gemma 4, and Granite, respectively.
  }
  \label{fig:token-length-kimi}
\end{figure*}

\begin{figure*}[p]
  \centering
  \includegraphics[width=\textwidth]
  {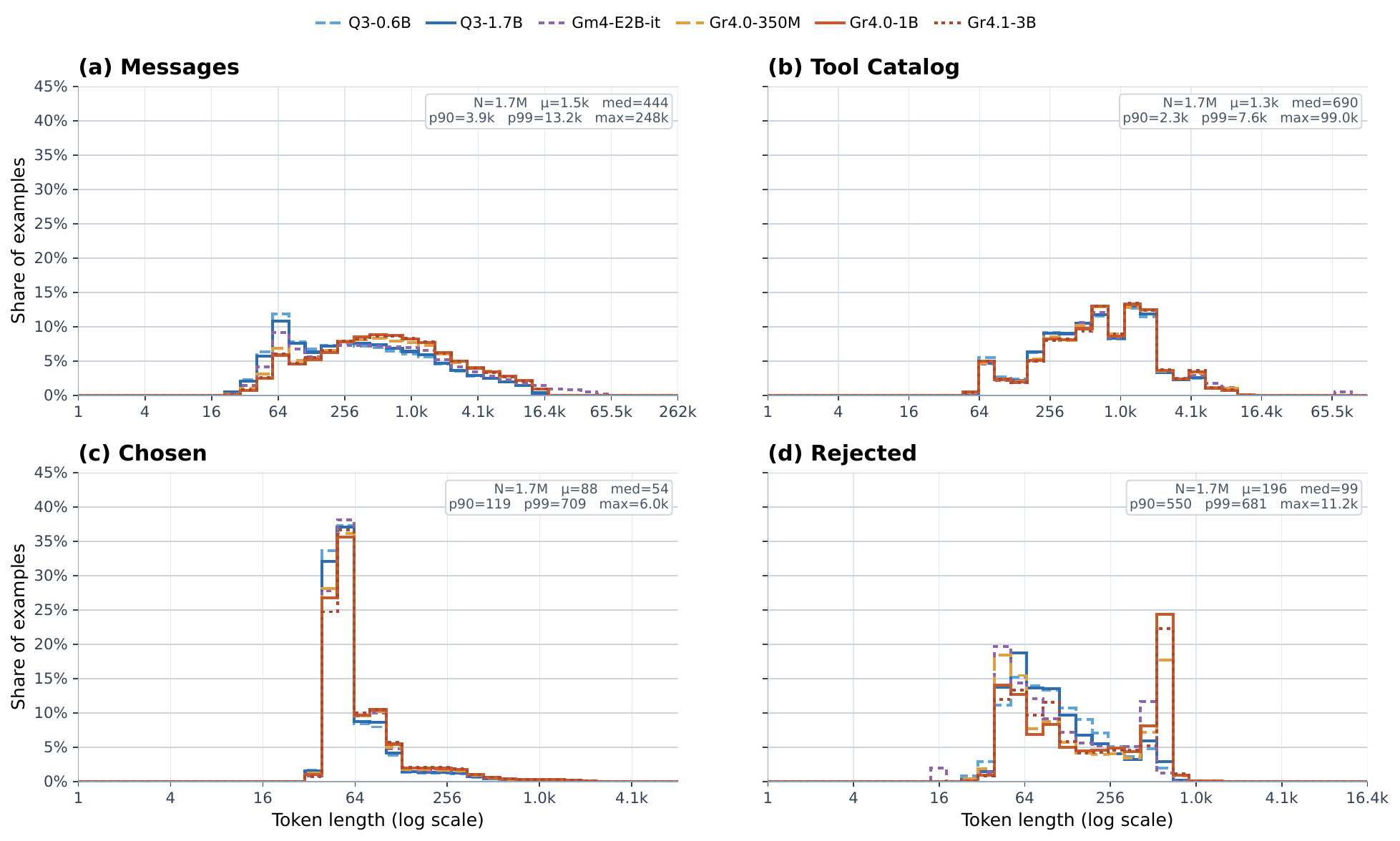}
  \caption{
    Token-length distributions for the OSS source partition, measured with the Granite 4.0 tokenizer. Panels show the conversation history (messages), the tool catalogue, and the chosen and rejected responses; interaction categories are pooled, and each curve corresponds to one generator subset, normalised independently. In-panel statistics summarise the component over the full partition. The horizontal axis is logarithmic. Q3, Gm4, and Gr denote Qwen3, Gemma 4, and Granite, respectively.
  }
  \label{fig:token-length-oss}
\end{figure*}

\begin{figure*}[p]
  \centering
  \includegraphics[width=\textwidth]
  {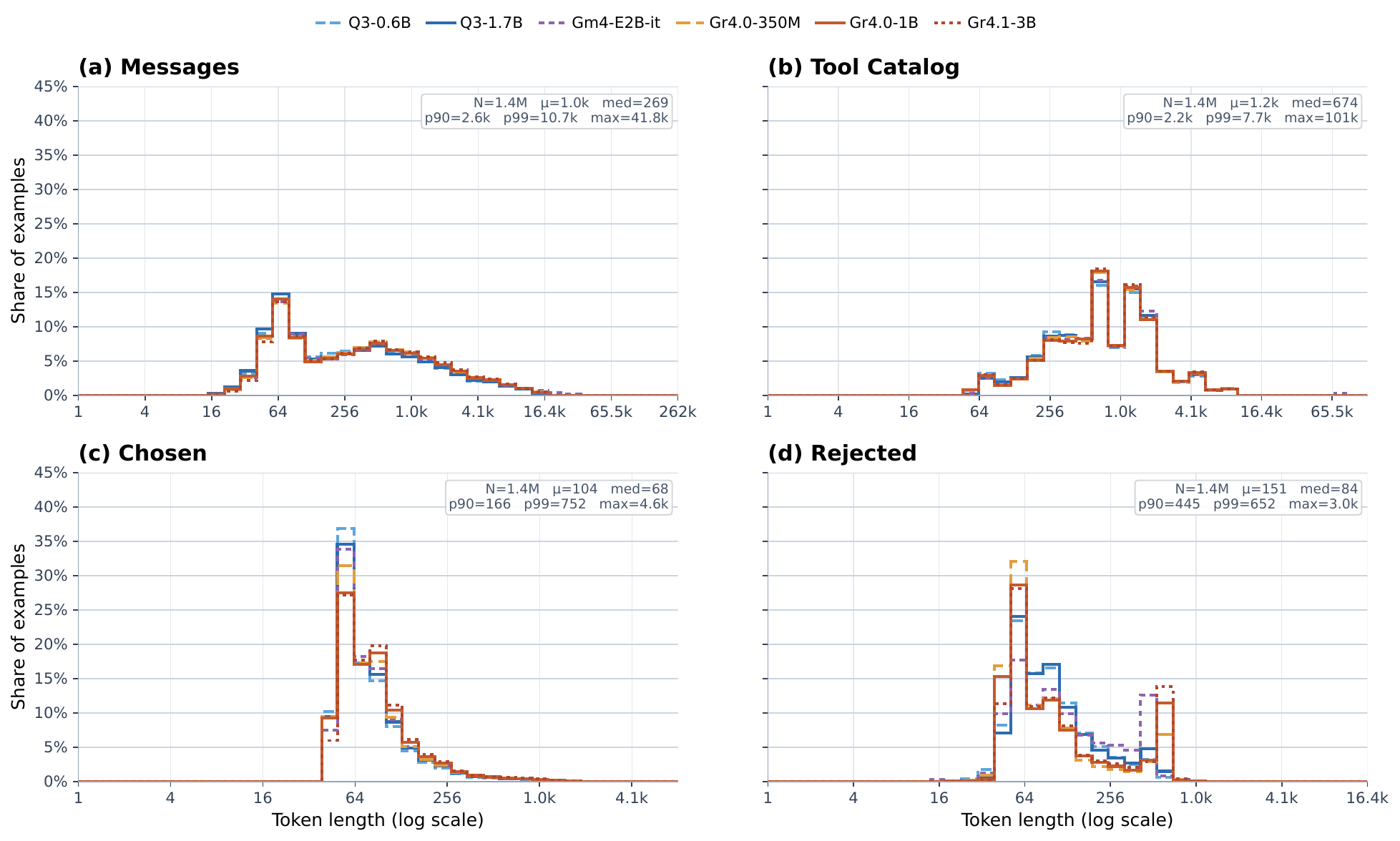}
  \caption{
    Token-length distributions for the Qwen source partition, measured with the Granite 4.0 tokenizer. Panels show the conversation history (messages), the tool catalogue, and the chosen and rejected responses; interaction categories are pooled, and each curve corresponds to one generator subset, normalised independently. In-panel statistics summarise the component over the full partition. The horizontal axis is logarithmic. Q3, Gm4, and Gr denote Qwen3, Gemma 4, and Granite, respectively.
  }
  \label{fig:token-length-qwen}
\end{figure*}

In addition to the ToolRL reward distributions, we characterise the sequence
lengths occurring in the generated preference data.
Figures~\ref{fig:token-length-kimi}--\ref{fig:token-length-qwen} show the
token-length distributions for the Kimi-K2, OSS, and Qwen source partitions,
respectively.

Each figure separates four components of a preference pair: the conversation history (messages), the tool catalogue, and the chosen and rejected responses. The four interaction categories (MT--MC, MT--SC, ST--MC, and ST--SC) are pooled, each curve corresponds to one generator subset and is normalised independently, and the in-panel statistics (count, mean, median, 90th and 99th percentiles, maximum) summarise the component over the full partition. All lengths are measured with the Granite 4.0 tokenizer, independently of the generator model, so the curves are directly comparable across generator subsets. Message and tool-catalogue lengths are nearly identical across generator subsets, as they stem from the same source conversations and differ only through subset membership; the chosen- and rejected-response panels characterise the output lengths of the preference pairs themselves. Token lengths span several orders of magnitude, so the horizontal axis is logarithmic.


\clearpage


\clearpage


\clearpage

%
%
\section{TOUMAII-ACE}
\label{app:toolace}

\begin{figure*}[t]
  \centering
  \includegraphics[width=\textwidth]
  {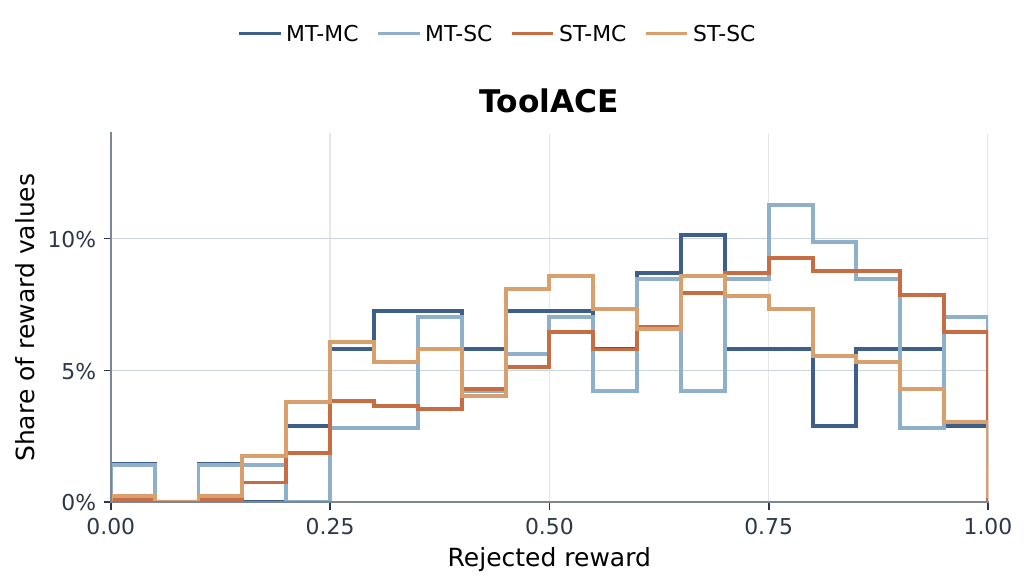}
  \caption{
    Distribution of rejected-response ToolRL rewards for TOUMAII-ACE, pooled
    across generator models. The four curves correspond to MT--MC, MT--SC,
    ST--MC, and ST--SC interactions. The distributions are unweighted and
    non-cumulative, use a bin width of \(0.05\), and are independently
    normalised. Chosen-response rewards are omitted because they are
    uniformly \(1.0\).
  }
  \label{fig:reward-aggregated-toolace}
\end{figure*}

\begin{figure*}[p]
  \centering
  \includegraphics[width=\textwidth]
  {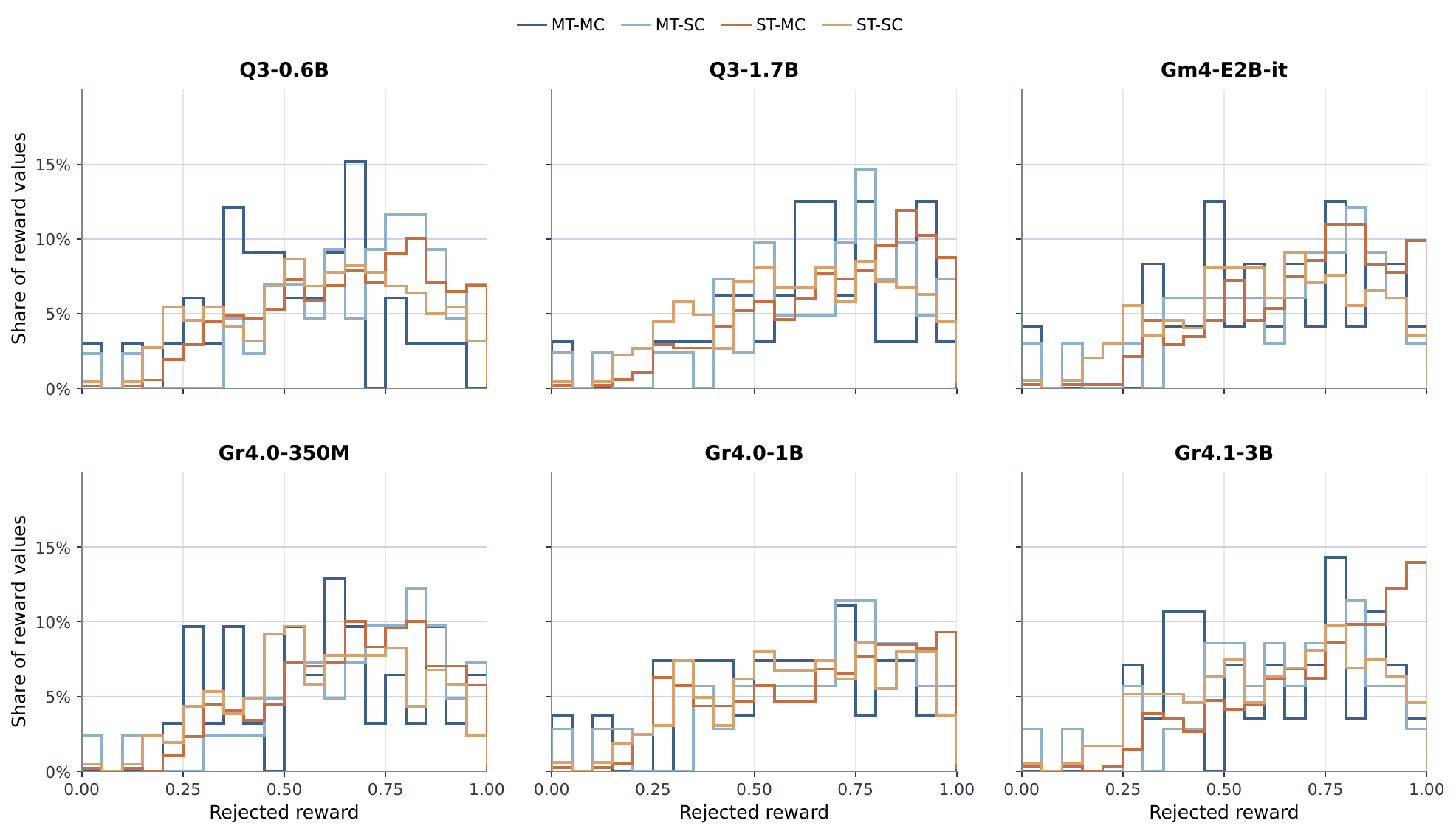}
    \caption{
    Model-specific distributions of rejected-response ToolRL rewards for
    TOUMAII-ACE. Each panel corresponds to one generator model, while the
    four lines represent MT--MC, MT--SC, ST--MC, and ST--SC interactions.
    The distributions are unweighted and non-cumulative, use a bin width of
    \(0.05\), and are independently normalised. Q3, Gm4, and Gr denote
    Qwen3, Gemma 4, and Granite, respectively. Chosen-response rewards are
    omitted because they are uniformly \(1.0\).
  }
  \label{fig:reward-by-model-toolace}
\end{figure*}

\begin{figure*}[p]
  \centering
  \includegraphics[width=\textwidth]
  {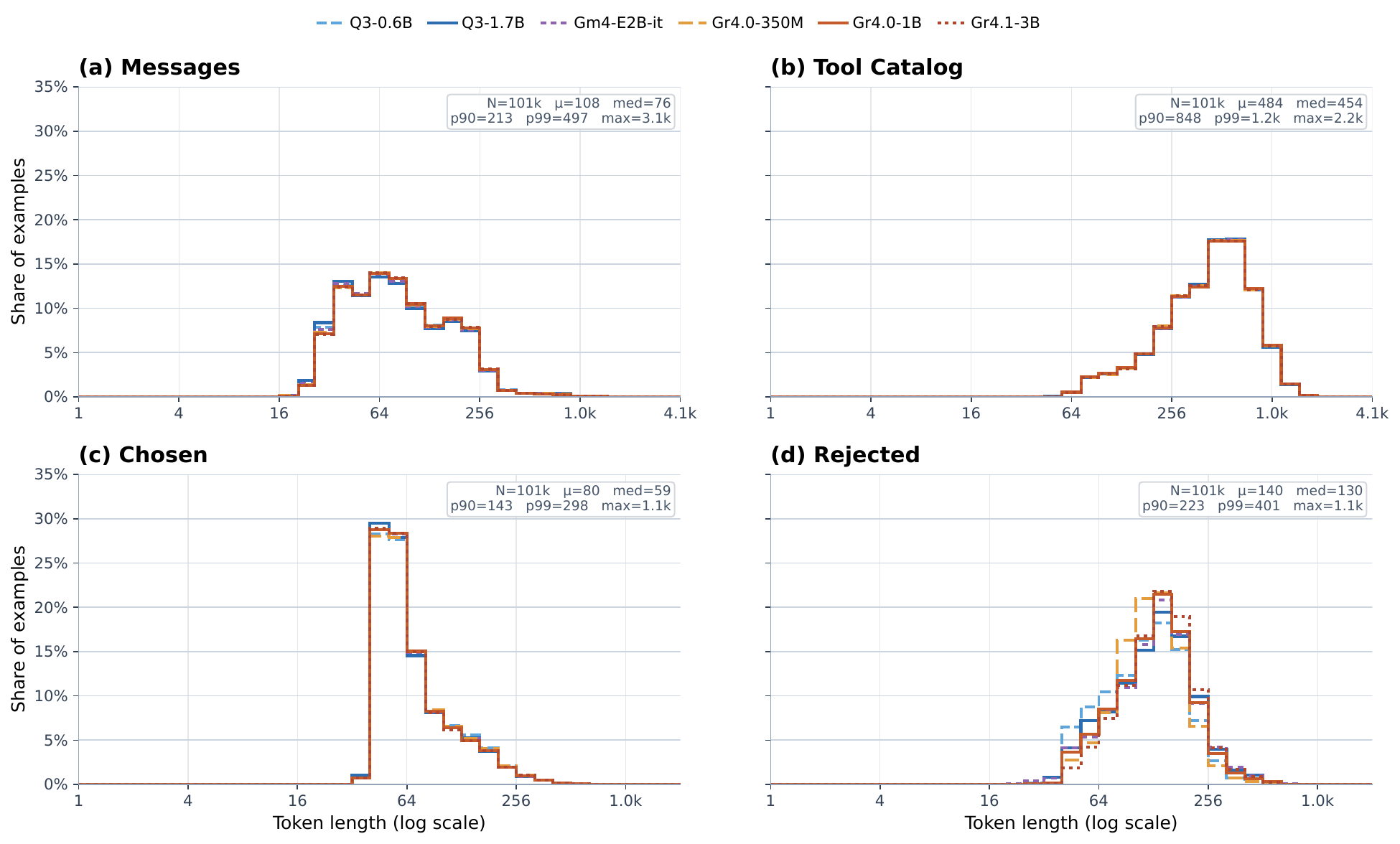}
  \caption{
    Token-length distributions for TOUMAII-ACE, measured with the Granite 4.0
    tokenizer. Panels show the conversation history (messages), the tool
    catalogue, and the chosen and rejected responses; interaction categories
    are pooled, and each curve corresponds to one generator subset,
    normalised independently. In-panel statistics summarise the component
    over the full dataset. The horizontal axis is logarithmic. Q3, Gm4, and
    Gr denote Qwen3, Gemma 4, and Granite, respectively.
  }
  \label{fig:token-length-toolace}
\end{figure*}

TOUMAII-ACE is constructed from ToolACE and follows the same
post-processing procedure as the Toucan-derived partitions.
Table~\ref{tab:toolace-composition} reports the resulting dataset
composition across generator models and interaction categories, while
Table~\ref{tab:toolace-funnel} summarises the post-processing funnel.
Figures~\ref{fig:reward-aggregated-toolace} and~\ref{fig:reward-by-model-toolace} report the rejected-response ToolRL reward distributions, and Figure~\ref{fig:token-length-toolace} the token-length distributions.

\begin{table}[!htbp]
  \caption{Exact number of preference pairs in TOUMAII-ACE per generator model and interaction category after post-processing and cross-subset de-duplication.}
  \label{tab:toolace-composition}
  \centering
  \small
  \setlength{\tabcolsep}{3pt}
  \begin{tabular}{lrrrrr}
    \toprule
    Generator model & MT--MC & MT--SC & ST--MC & ST--SC & Total \\
    \midrule
    Qwen3-0.6B       & 50 & 405 & 3,097 & 13,965 & 17,517 \\
    Qwen3-1.7B       & 44 & 416 & 2,815 & 14,039 & 17,314 \\
    Gemma-4-E2B-it   & 34 & 346 & 2,731 & 13,927 & 17,038 \\
    Granite-4.0-350M & 40 & 327 & 2,751 & 13,781 & 16,899 \\
    Granite-4.0-1B   & 38 & 245 & 2,393 & 13,178 & 15,854 \\
    Granite-4.1-3B   & 32 & 274 & 2,382 & 13,536 & 16,224 \\
    \midrule
    All models        & 238 & 2,013 & 16,169 & 82,426 & 100,846 \\
    \bottomrule
  \end{tabular}
\end{table}

\begin{table}[!htbp]
  \caption{Post-processing funnel for TOUMAII-ACE. A row can violate several validity rules, so per-rule counts can sum to more than the number of unique rows removed. Cross-subset duplicates are multi-turn rows whose (prompt, chosen, rejected) triplet also occurs in the corresponding single-turn subsets. All 56 such duplicates were in MT--SC, and post-clean checks found no remaining overlaps.}
  \label{tab:toolace-funnel}
  \centering
  \small
  \begin{tabular}{lr}
    \toprule
    & ToolACE \\
    \midrule
    Generated candidate pairs & 130,782 \\
    \midrule
    \quad reference-identical pair (zero reward difference) & 29,605 \\
    \quad long free-text value in chosen response ($>30$ words) & 264 \\
    \quad rejected response equals a later chosen turn & 288 \\
    Rows removed as invalid tool errors & 29,880 \\
    Cross-subset duplicates (multi-turn vs.\ single-turn) & 56 \\
    \midrule
    Final preference pairs (after de-duplication) & 100,846 \\
    Retention & 77.1\% \\
    \bottomrule
  \end{tabular}
\end{table}

%
%

\clearpage

\clearpage

%
%

\clearpage

\section{Training details}
\label{app:training-details}

\paragraph{Data preparation.}
The sampling and conversion protocol is summarised in
Section~\ref{sec:setup}; we record here the details omitted there.
The ToolRM corpus is \textit{ibm-research/ToolRM-train-data}.
Sequence lengths are measured after appending the EOS token. Both rendered
datasets are deterministically shuffled with seed 42 and subsequently
partitioned into non-overlapping 50{,}000-example slices. Each slice is used
in an independent training run initialised from the same Granite checkpoint
and trained with the fixed seed. Consequently,
variation across runs reflects differences in training-data membership rather
than training-seed variance.

\paragraph{Hyperparameters.}
Table~\ref{tab:hyperparams} lists the complete LoRA and DPO configuration.

\begin{table}[h]
  \caption{LoRA and DPO training configuration.}
  \label{tab:hyperparams}
  \centering
  \small
  \begin{tabular}{ll}
    \toprule
    Base model & \texttt{ibm-granite/granite-4.1-3b} \\
    LoRA rank / $\alpha$ / dropout & 64 / 64 / 0.0 \\
    LoRA target modules & Attention and MLP projections (\texttt{q/k/v/o}, \texttt{gate/up/down}) \\
    DPO $\beta$ & 0.1 \\
    Optimiser / learning rate / schedule &
    AdamW (fused) / $1\times10^{-6}$ / linear with 5\% warmup \\
    Effective batch size & 64 ($2 \times 32 \times 1$) \\
    Epochs per slice & 1 \\
    Max sequence length & 2048 tokens \\
    Precision / hardware & BF16, 1$\times$ AMD MI300X \\
    Seed & 42 \\
    \bottomrule
  \end{tabular}
\end{table}

\section{Complete Evaluation Results}
\label{app:complete-results}

Table~\ref{tab:full-results} reports the complete BFCL~v4 results for
the Granite baseline and each 50{,}000-example training-data slice. Slice identifiers denote non-overlapping partitions of the deterministically shuffled training corpora, rather than different random seeds. Web-search accuracy is excluded from the evaluation due to external API requirements. For completeness, we mention web-search accuracy in the table. 

\begin{table*}[t]
  \caption{
    Complete BFCL~v4 results (\%). ``Base'' denotes the unmodified
    \texttt{ibm-granite/granite-4.1-3b} checkpoint. Irrelevance detection
    is the unweighted mean of the non-live and live irrelevance categories.
  }
  \label{tab:full-results}
  \centering
  \scriptsize
  \setlength{\tabcolsep}{3pt}
  \renewcommand{\arraystretch}{0.95}
  \begin{tabular}{@{}p{0.25\textwidth}rrrrrrrrr@{}}
    \toprule
    & Base
    & \multicolumn{4}{c}{ToolRM slice}
    & \multicolumn{4}{c}{TOUMAII slice} \\
    \cmidrule(lr){3-6}
    \cmidrule(lr){7-10}
    Metric & -- & 000 & 001 & 002 & 003 & 000 & 001 & 002 & 003 \\
    \midrule

    \multicolumn{10}{@{}l}{\textit{Overall}} \\
    Overall accuracy
      & 37.76 & 36.57 & 35.78 & 36.20 & 35.63
      & 37.58 & 37.26 & 36.62 & 36.58 \\

    \addlinespace
    \multicolumn{10}{@{}l}{\textit{Non-live}} \\
    Non-live AST
      & 87.04 & 86.38 & 86.48 & 86.10 & 86.25
      & 86.85 & 86.81 & 86.58 & 86.98 \\
    \quad Simple AST
      & 75.67 & 74.50 & 73.42 & 74.42 & 74.50
      & 74.42 & 74.75 & 73.83 & 74.42 \\
    \quad Multiple
      & 93.00 & 94.00 & 94.00 & 94.00 & 94.50
      & 93.00 & 93.00 & 93.50 & 93.50 \\
    \quad Parallel
      & 92.00 & 90.50 & 91.50 & 91.00 & 90.50
      & 92.00 & 91.50 & 91.50 & 92.00 \\
    \quad Parallel multiple
      & 87.50 & 86.50 & 87.00 & 85.00 & 85.50
      & 88.00 & 88.00 & 87.50 & 88.00 \\

    \addlinespace
    \multicolumn{10}{@{}l}{\textit{Live}} \\
    Live accuracy
      & 76.54 & 75.13 & 75.13 & 75.35 & 75.20
      & 77.79 & 77.42 & 77.57 & 77.42 \\
    \quad Simple
      & 77.91 & 74.42 & 75.97 & 74.81 & 75.58
      & 80.23 & 80.23 & 78.68 & 78.68 \\
    \quad Multiple
      & 76.54 & 75.50 & 75.12 & 75.69 & 75.31
      & 77.40 & 76.92 & 77.40 & 77.21 \\
    \quad Parallel
      & 75.00 & 75.00 & 75.00 & 75.00 & 75.00
      & 75.00 & 75.00 & 75.00 & 75.00 \\
    \quad Parallel multiple
      & 62.50 & 66.67 & 66.67 & 66.67 & 66.67
      & 70.83 & 70.83 & 75.00 & 75.00 \\

    \addlinespace
    \multicolumn{10}{@{}l}{\textit{Multi-turn}} \\
    Multi-turn accuracy
      & 23.00 & 21.38 & 20.00 & 21.25 & 21.50
      & 21.88 & 21.38 & 20.62 & 20.88 \\
    \quad Base
      & 29.50 & 27.00 & 26.00 & 26.00 & 28.50
      & 28.50 & 27.00 & 28.50 & 30.50 \\
    \quad Missing function
      & 20.00 & 22.00 & 19.00 & 23.00 & 22.00
      & 17.50 & 16.00 & 15.00 & 14.50 \\
    \quad Missing parameter
      & 18.00 & 15.00 & 14.50 & 16.00 & 15.00
      & 17.50 & 18.50 & 16.50 & 14.50 \\
    \quad Long context
      & 24.50 & 21.50 & 20.50 & 20.00 & 20.50
      & 24.00 & 24.00 & 22.50 & 24.00 \\

    \addlinespace
    \multicolumn{10}{@{}l}{\textit{Agentic}} \\
    Web-search accuracy
      & 1.00 & 1.50 & 1.00 & 1.00 & 1.00
      & 2.50 & 2.00 & 3.00 & 3.00 \\
    \quad With snippet
      & 1.00 & 2.00 & 1.00 & 1.00 & 1.00
      & 3.00 & 2.00 & 3.00 & 3.00 \\
    \quad Without snippet
      & 1.00 & 1.00 & 1.00 & 1.00 & 1.00
      & 2.00 & 2.00 & 3.00 & 3.00 \\
    Memory accuracy
      & 33.55 & 29.89 & 28.17 & 28.60 & 26.67
      & 35.70 & 35.48 & 34.19 & 33.33 \\
    \quad Key--value
      & 12.26 & 9.03 & 9.03 & 9.68 & 10.97
      & 14.19 & 20.00 & 13.55 & 14.84 \\
    \quad Vector
      & 45.16 & 36.77 & 35.48 & 36.77 & 33.55
      & 50.32 & 46.45 & 45.16 & 41.94 \\
    \quad Recursive summarisation
      & 43.23 & 43.87 & 40.00 & 39.35 & 35.48
      & 42.58 & 40.00 & 43.87 & 43.23 \\

    \addlinespace
    \multicolumn{10}{@{}l}{\textit{Detection aggregates}} \\
    Relevance detection
      & 81.25 & 81.25 & 81.25 & 81.25 & 81.25
      & 87.50 & 87.50 & 87.50 & 87.50 \\
    Irrelevance detection
      & 75.97 & 77.24 & 77.89 & 77.64 & 75.03
      & 69.11 & 69.23 & 65.75 & 66.08 \\

    \bottomrule
  \end{tabular}
\end{table*}

\clearpage

\section{Extended Limitations and Responsible Release}
\label{app:limitations}

\subsection{Reference and Validation limitations}
TOUMAII is constructed by pairing reference responses from quality-filtered trajectories (if quality indicators are present) with model-generated rejected responses. Our automated validation measures agreement with the reference tool-call (in this case, this is the chosen/completion response) rather than any execution-level or semantic correctness. Consequently, disagreement with the reference does not necessarily equate to a rejected response being invalid in every setting, and some retained pairs may contain alternative valid calls or reflect errors in the reference trajectory. Certain cleaning steps help prevent this situation to some extent, although, due to the nature of the generation, complete adherence cannot be guaranteed. This is particularly relevant for multi-turn examples, where correctness is established at the trajectory level, and an intermediate reference call may later be corrected after tool feedback.
\subsection{Inherited source and generator limitations}
The dataset inherits the features and limitations of the source datasets, including Toucan and ToolACE, in addition to the language models used to generate rejected responses. The generator pool comprises models which are publicly released instruction-tuned models with their own safety mechanisms, and we do not independently evaluate those mechanisms. As with any model generation, risks of hallucination, bias, and undesirable outputs can be expected in the generator pipeline, and necessary downstream validations should still be in place.
\subsection{Potential downstream risks}
TOUMAII is intended for training and evaluating tool-calling models. Improved tool-call reliability can reduce malformed calls and incorrect API interactions, but models trained on the dataset may still produce erroneous or inappropriate actions. Such errors may be more consequential when models are connected to external systems such as databases, APIs, or enterprise software. The dataset should therefore not be interpreted as certifying the safety or correctness of downstream models, and application-specific validation, permission controls, and other deployment safeguards remain necessary for consequential tool use.
\subsection{Licensing and provenance}
TOUMAII is derived from existing tool-calling datasets and augments them with newly generated rejected responses and annotations. The original source datasets and models are credited in the paper, and the corresponding licenses and usage terms should be retained in the released dataset documentation. The derived dataset should state its own license separately and clearly identify the provenance of its source examples.


\end{document}